# NFRsTDO v1.2's Terms, Properties, and Relationships -- A Top-Domain Non-Functional Requirements Ontology

**Luis Olsina, María Fernanda Papa,** and **Pablo Becker**

GIDIS_Web, Facultad de Ingeniería, UNLPam, General Pico, LP, Argentina

[olsinal, pmfer, beckerp]@ing.unlpam.edu.ar

**Abstract.** This preprint specifies and defines all the Terms, Properties, and Relationships of NFRsTDO (*Non-Functional Requirements Top-Domain Ontology*). NFRsTDO v1.2, whose UML conceptualization is shown in Figure 1 is a slightly updated version of its predecessor, namely NFRsTDO v1.1. NFRsTDO is an ontology mainly devoted to quality (non-functional) requirements and quality/cost views, which is placed at the top-domain level in the context of a multilayer ontological architecture called FCD-OntoArch (*Foundational, Core, Domain, and instance Ontological Architecture for sciences*). Figure 2 depicts its five tiers, which entail Foundational, Core, Top-Domain, Low-Domain, and Instance. Each level is populated with ontological components or, in other words, ontologies. Ontologies at the same level can be related to each other, except at the foundational level, where only ThingFO (*Thing Foundational Ontology*) is found. In addition, ontologies' terms and relationships at lower levels can be semantically enriched by ontologies' terms and relationships from the higher levels. NFRsTDO's terms and relationships are mainly extended/reused from ThingFO, SituationCO (*Situation Core Ontology*), ProcessCO (*Process Core Ontology*), and FRsTDO (*Functional Requirements Top-Domain Ontology*). Stereotypes are the used mechanism for enriching NFRsTDO terms. Note that annotations of updates from the previous version (NFRsTDO v1.1) to the current one (v1.2) can be found in Appendix A.

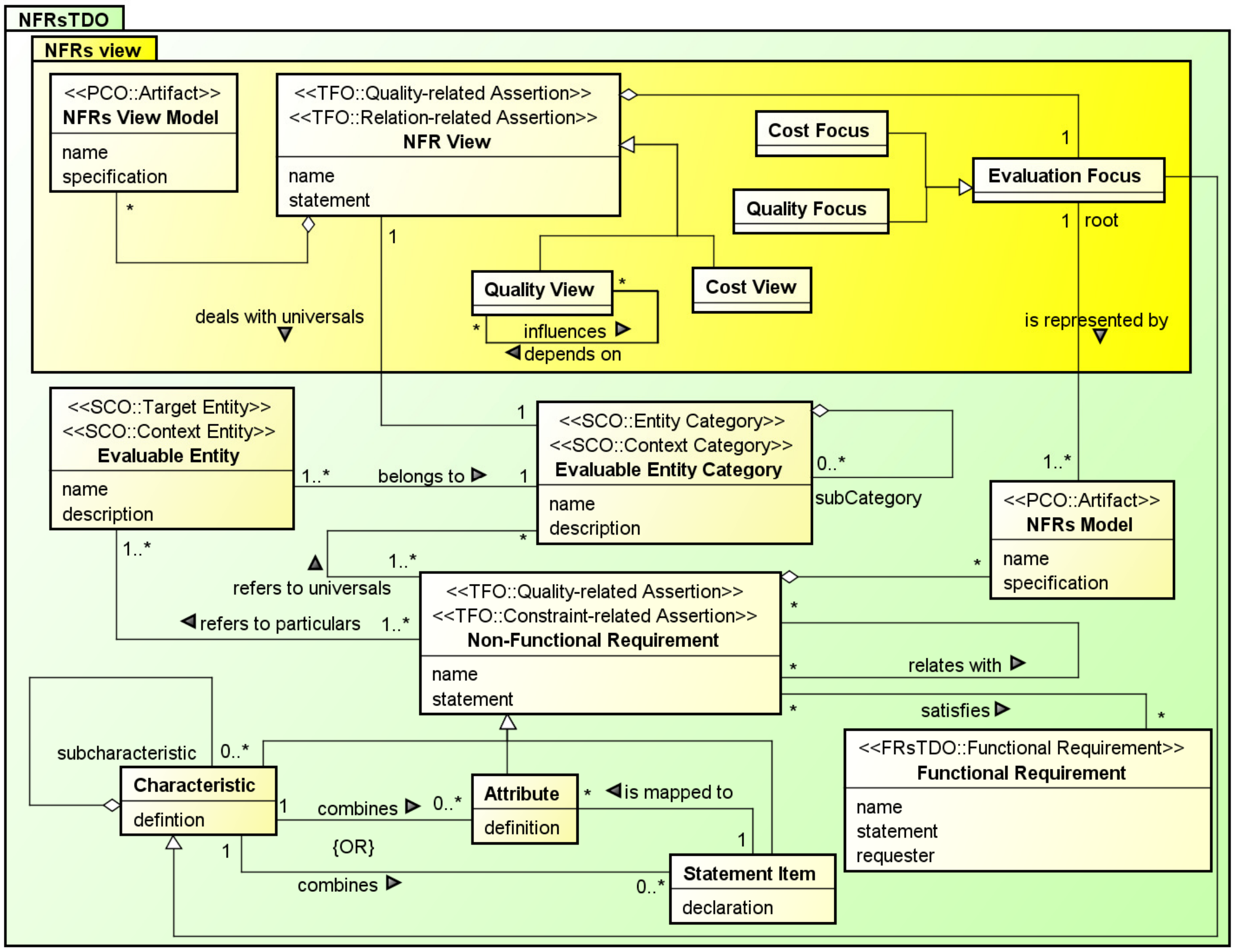

**Figure 1.** NFRsTDO v1.2: Top-Domain Ontology for Non-Functional Requirements, which is placed at the higher-domain level of FCD-OntoArch (see Figure 2). Note this is a revised version of NFRsTDO v1.1 (named in [1, 11] as the NFRs ontology). Annotations of updates from the previous version (v1.1) to the current one can be found in Appendix A. Also, note that TFO stands for ThingFO [6, 7], SCO for SituationCO [8], FRsTDO for Functional Requirements TDO [13], and PCO for ProcessCO [2, 3].

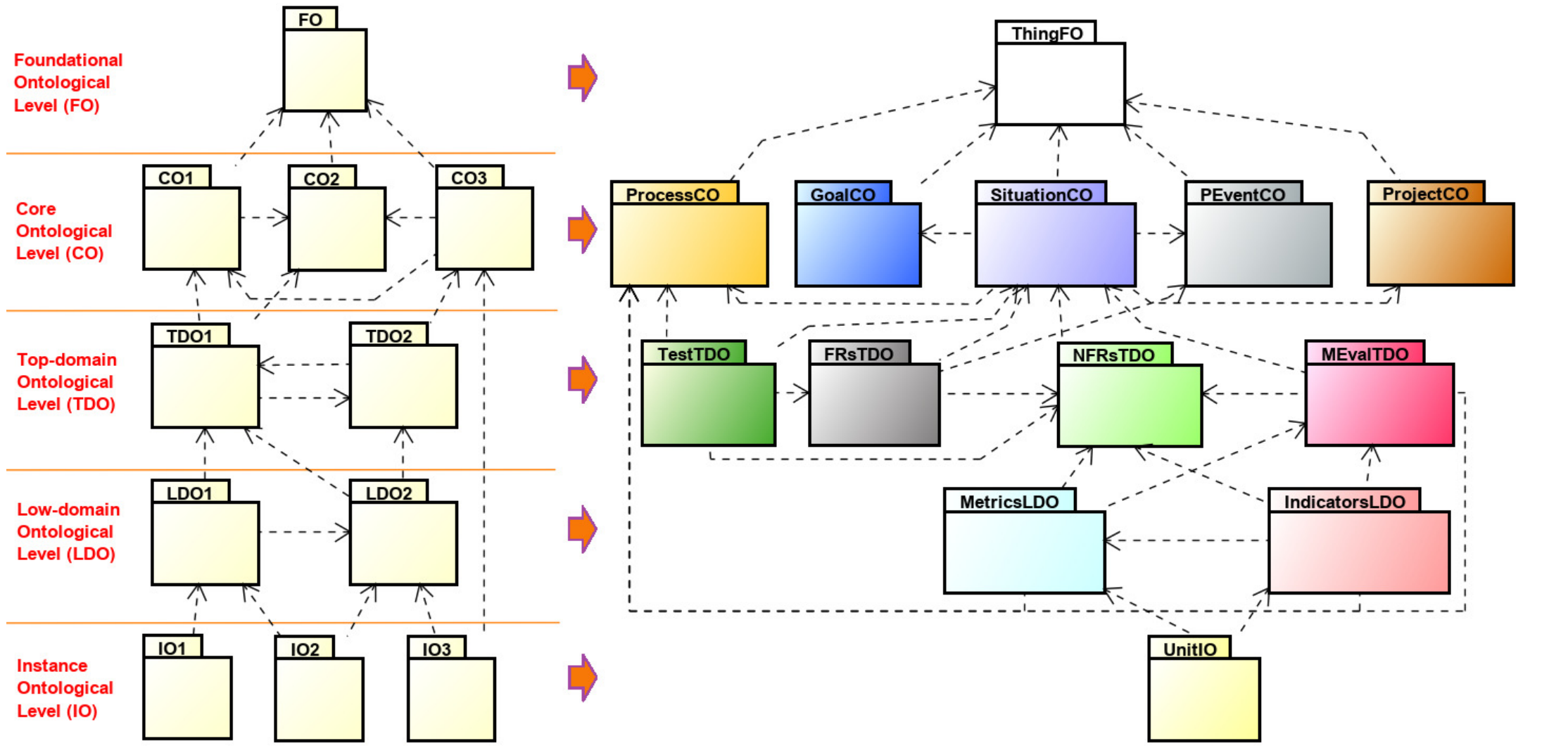

**Figure 2.** It shows the allocation of the NFRsTDO component or module in the context of the five-tier ontological architecture so-called FCD-OntoArch (*Foundational, Core, Domain, and instance Ontological Architecture for sciences*) [6].

## Non-Functional Requirements Component – NFRsTDO v1.2's Terms

| Term | Definition |
|---|---|
| **Attribute**<br>(synonym: **Property**; **Elementary Aspect**) | It is an NFR that represents a measurable and evaluable physical or abstract aspect attributed to a particular entity or its category by a human agent.<br><br>Note 1: The aspect attributed to a particular thing can deal with its properties, its powers, or a combination of both, as represented in ThingFO [6].<br><br>Note 2: An Attribute is a measurable and evaluable elementary NFR, i.e., an elementary quality to be quantified.<br><br>Note 3: An Attribute can be quantified by employing metrics [9] and interpreted utilizing elementary indicators [10]. |
| **Characteristic**<br>(synonym: **Dimension**; **Factor**; **Non-elementary Aspect**; **Calculable Concept**; **Evaluable Concept**) | It is an NFR that represents an evaluable, non-elementary aspect attributed to a particular entity or its category by a human agent.<br><br>Note 1: A characteristic can be evaluated but cannot be measured as an Attribute –at least in a non-very trivial way such as "good" or "bad". Instead, a Characteristic is a non-elementary NFR, which combines Attributes or Statement Items.<br><br>Note 2: A Characteristic can have sub-characteristics.<br><br>Note 3: A Characteristic can be calculated and interpreted utilizing derived indicators and aggregation functions [10]. |
| **Evaluable Entity**<br>(synonym: **Evaluable Particular Entity**; **Evaluable Particular** | It is a Target Entity or Context Entity that represents a particular (concrete) entity to be evaluated. |
| **Object**) | Note 1: Depending on a given situation (SituationCO in [8]), an Evaluable Entity has the semantics of Target Entity or Context Entity, which in turn are things. The term Thing is defined in ThingFO [6].<br><br>Note 2: Examples of Evaluable Entities are particular (concrete) products (model, software source code, document, smartphone device, building, etc.), systems, resources, work processes, services, and infrastructure, among many others. |
| **Evaluable Entity Category** (synonym: **Evaluable Universal**) | It is an Entity Category or Context Category to which particular Evaluable Entities belong to.<br><br>Note 1: Depending on a given situation (SituationCO in [8]), an Evaluable Entity Category has the semantics of Entity Category or Context Category, which in turn have the semantics of Thing Category from the ThingFO ontology [7].<br><br>Note 2: A universal (or category) can be classified into Evaluable Entity Category, Developable Entity Category [13], Testable Entity Category [14, 15], Observable Entity Category, etc., depending on the intention of the human agent.<br><br>Note 3: Examples of Evaluable Entity Categories can be Software Product Category, Service Category, Software System Category, Resource Category, and Work Process Category, among others. |
| **Functional Requirement (FR)** | It is an Assertion on Particulars that specifies what the existing Developable Entity (or sub-Developable Entity) does, or the new one shall do by referring to its features and/or capabilities considering a given requester's need.<br><br>Note 1: A Functional Requirement has the semantics of Assertion on Particulars from the ThingFO ontology [2], and more specifically it has the semantics of one or more assertions such as Structure-, Behavior-, Action-, Relation-related Assertions [7] depending on the particular situation.<br><br>Note 2: A Functional Requirement (term reused from the FRsTDO ontology [13]) can require to be satisfied by constraints or 'ilities', which are stated by Non-Functional Requirements. |
| **Non-Functional Requirement (NFR)** | It is a Quality-, Constraint-related Assertion that specifies an aspect in the form of a Characteristic, Attribute, or Statement Item to be evaluated on how or how well an Evaluable Entity performs or shall perform.<br><br>Note: A Non-Functional Requirement is often referred to as an 'ility'. |
| **NFRs Model** (synonym: **Quality Model**) | It is an Artifact that specifies and represents Non-Functional Requirements.<br><br>Note 1: A NFRs Model has the semantics of Artifact, |
| | which is a term coming from ProcessCO [2].<br><br>Note 2: For example, the structure of the NFRs Model for quality represented in the ISO 25010 standard [5], hierarchically models a set of Characteristics, sub-characteristics, and Attributes (or properties as regards the used term in it), as well as the relationships between them, which provide the basis for specifying the NFRs and their further evaluation for software products and systems. |
| **Statement Item**<br>(synonym:<br>**Item; Guideline; Heuristic**) | It is an NFR that represents a declared textual expression of an evaluable physical or abstract aspect asserted for a particular entity or its category by a human agent.<br><br>Note 1: A Statement Item can represent for instance an assertive element (item) in a questionnaire instrument, a heuristic checklist, or a coding style guide.<br><br>Note 2: A Statement Item can be quantified by using the scales designed in a questionnaire instrument, or utilizing other qualitative ways. Also, it can be mapped to Attributes. |
| **NFRs View sub-ontology's Terms that are included in the NFRs component** | |
| **Cost Focus** | It is an Evaluation Focus for cost. |
| **Cost View**<br>(synonym: **Cost Perspective**) | It is an NFR View for cost. |
| **Evaluation Focus** | It is a Characteristic that represents the root of a NFRs Model. |
| **NFR View**<br>(synonym: **NFR Perspective**) | It is an Assertion on Universals that relates one Evaluable Entity Category with one Evaluation Focus.<br><br>Note 1: A NFR View has the semantics of Assertion on Universals from the ThingFO ontology [7], and more specifically, it has the semantics of Quality-related Assertion and Relation-related Assertion.<br><br>Note 2: The Evaluable Entity Category must be the super-category, i.e., the highest abstraction level of an Entity Category of value to be evaluated in a given organization. For example, names of entity super-categories of interest to be evaluated are Resource Category, Process Category, Software Product Category, Hardware Product Category, System Category, Service Category, System-in-use Category, among others.<br><br>Note 3: The Evaluable Focus must be the root Characteristic, which represents the Characteristic with the highest level of abstraction in a Non-Functional Requirements (NFRs) Model of value to be evaluated in a given organization. For example, names of root Characteristics of interest to be evaluated are Resource |
| | Quality, Process Quality, Internal Quality, External Quality, Quality in Use, and Service Quality, among others.<br><br>Note 4: Names of NFR Views are: Quality View and Cost View [12]. |
| **NFRs View Model** | It is an Artifact that specifies and represents NFRs Views.<br><br>Note 1: A NFRs View Model has the semantic of Artifact, which is a term coming from ProcessCO [2].<br><br>Note 2: An example of instantiation of NFRs Views for quality is modeled in Figure 3, in [12]. |
| **Quality Focus** | It is an Evaluation Focus for quality.<br><br>Note: Names of Quality Focuses are, for example, Process Quality, Internal Quality, External Quality, and Quality in Use, among others. |
| **Quality View**<br>(synonym: **Quality Perspective**) | It is an NFR View for quality.<br><br>Note: Names of Quality Views are, for example, Resource Quality View, Process Quality View, and Software Product Quality View, among others. In [12] authors say: "*the Resource Quality View influences the Process Quality View. For example, if a development team uses a new tool or method –both considered as entities of the Resource Entity super-category- this fact impacts directly in the quality of the development process they are carrying out. Likewise, the Process Quality View influences the Software Product Quality View. The Product Quality View influences the System Quality, and this in turn influences the System-in-Use Quality View. Conversely, the depends on relationship has the opposite semantic.*" |

*Amount of Own or Reused/Extended Terms: 15*

## Non-Functional Requirements Component – NFRsTDO v1.2's Attributes or Properties

| Term | Attribute | Definition |
|---|---|---|
| **Attribute** | **definition** | An unambiguous textual meaning of the Attribute, which as an Assertion refers to an elementary aspect of an Evaluable Entity or Evaluable Entity Category. |
| **Characteristic** | **definition** | An unambiguous textual meaning of the Characteristic, which as an Assertion refers to a non-elementary aspect of an Evaluable Entity or Evaluable Entity Category. |
| **Evaluable Entity** | **name** | Label or name that identifies the Evaluable Entity. |
| | **description** | An unambiguous textual statement describing the Evaluable Entity. |
| **Evaluable Entity Category** | **name** | Label or name that identifies the Evaluable Entity Category. |
| | **description** | An unambiguous textual description of the aim of the Evaluable Entity Category as universal. |
| **Functional Requirement** | **name** | Label or name that identifies the Functional Requirement. |
| | **statement** | An explicit declaration of what the existing Developable Entity does, or the new one shall do. |
| | **requester** | An agent that requires or establishes the Functional Requirement. |
| **NFRs Model** | **name** | Label or name that identifies the Non-Functional Requirements (NFRs) Model. |
| | **specification** | The explicit and detailed representation or model of Non-Functional Requirements in a given language. |
| **Non-Functional Requirement** | **name** | Label or name that identifies the Non-Functional Requirement. |
| | **statement** | An explicit declaration of the Characteristic, Attribute, or Statement Item to be evaluated on how or how well an Evaluable Entity or Evaluable Entity Category performs or shall perform. |
| **NFR View** | **name** | Label or name that identifies the Non-Functional Requirement View. |
| | **statement** | An explicit declaration of the one-to-one relationship between an Evaluable Entity Category and its Evaluation Focus, which is linked by the NFR View concept. |
| **NFRs View Model** | **name** | Label or name that identifies the Non-Functional Requirements (NFRs) View Model. |
| | **specification** | The explicit and detailed representation or model of Non-Functional Requirements Views in a given language. |
| **Statement Item** | **declaration** | An unambiguous textual expression of the Statement Item, which as an Assertion refers to an elementary aspect of an Evaluable Entity or Evaluable Entity Category. |

*Amount of Attributes (Properties): 18*

## Non-Functional Requirements Component – NFRsTDO v1.2's Non-taxonomic Relationships

| Relationship | Definition |
|---|---|
| **belongs to** | Evaluable Entities belong to one Evaluable Entity Category. |
| **combines (x2)** | A Characteristic combines none or more Attributes. |
| | A Characteristic combines none or more Statement Items. |
| **deals with universals** | An NFR View deals with one Evaluable Entity Category as universal. |
| **depends on** | A Quality View depends on none or more Quality Views. |
| **influences** | A Quality View influences none or more Quality Views. |
| **is represented by** | An Evaluation Focus is represented by one or more Non-Functional Requirements (NFRs) Models. |
| **is mapped to** | A Statement Item is mapped to none or more Attributes. |
| **refers to particulars** | A Non-Functional Requirement refers to one or more Evaluable Entities as particulars. |
| **refers to universals** | A Non-Functional Requirement refers to none or more Evaluable Entity Categories as universals. |
| **relates with** | A Non-Functional Requirement relates with none or more Non-Functional Requirements. |
| **satisfies** | A Non-Functional Requirement satisfies none or more Functional Requirements. |

*Amount of non-taxonomic relationships: 12*

# Appendix A: Updates from NFRsTDO v1.1 to NFRsTDO v1.2

The main updates in NFRsTDO v1.2 w.r.t. v1.1 are listed below. Note that the NFRsTDO v1.0 ontology was simply named NFRs and is documented in [1] (and earlier in [11]).

- The term Non-Functional Requirement had the semantics of a mixture of Assertions, namely: Quality-related Assertion, Quantity-related Assertion, and Constraint-related Assertion. We have removed the Quantity-related Assertion in NFRsTDO v1.2.
- The term Evaluable Entity Category had the stereotype <<Thing Category>>. Now we have updated it considering Note 1 in the above definition of the term Evaluable Entity Category. This note states that "depending on a given situation (SituationCO in [8]), an Evaluable Entity Category has the semantics of Entity Category or Context Category, which in turn have the semantics of Thing Category from the ThingFO ontology [7]".
- The non-taxonomic reflexive relationship "relates with" was added to the term Non-Functional Requirement. In addition, the non-taxonomic relationship "is mapped to" was also added between the terms Statement Item and Attribute.
- We have included in the NFRsTDO v2.0 component the term Functional Requirement, which is fully reused from the FRsTDO ontology. Additionally, we have added the non-taxonomic relationship "satisfies" between the terms Non-Functional Requirement and Functional Requirement.
- The former double non-taxonomic relationship "refers to" was renamed as "refers to particulars" between the terms Non-Functional Requirement and Evaluable Entity, and as "refers to universals" between the terms Non-Functional Requirement and Evaluable Entity Category, respectively.
- The definitions of some terms that enrich NFRsTDO v1.2 were updated and harmonized according to the higher-level ontologies. This is because ThingFO, SituationCO, and ProcessCO were revised after NFRsTDO v1.1.
- Figure 1 was updated accordingly. Also, Figure 2 was updated to incorporate PEventCO [4], which is a new ontological component for particular discrete events we have developed.

## References

[1] Becker P., Tebes G., Peppino D., and Olsina L.: Applying an Improving Strategy that embeds Functional and Non-Functional Requirements Concepts, *Journal of Computer Science and Technology*, vol. 19, no. 2, pp. 153–175, (2019), doi: https://doi.org/10.24215/16666038.19.e15.

[2] Becker P. and Olsina L.: "ProcessCO v1.3's Terms, Properties, Relationships and Axioms - A Core Ontology for Processes", Preprint in Research Gate, August 2021, Available at https://www.researchgate.net/publication/353719307_ProcessCO_v13%27s_Terms_Properties_Relationships_and_Axioms_-_A_Core_Ontology_for_Processes, DOI: http://dx.doi.org/10.13140/RG.2.2.23196.82564. Also at https://doi.org/10.48550/arXiv.2108.02816.

[3] Becker P., Papa M.F., Tebes G., Olsina L.: Discussing the Applicability of a Process Core Ontology and Aspects of its Internal Quality. *Software Quality Journal*, Springer, 30:(4), 1003-1038, DOI: https://doi.org/10.1007/s11219-022-09592-3, (2022).

[4] Blas M. J., Gonnet S., Becker P., Olsina, L.: Ontología para la Representación de Entidades con Comportamientos basados en Eventos. A Core Ontology for the Representation of Entities with Event-based Behaviors. In: *Electronic Journal of SADIO, EJS*, 21:(2), pp. 17-41, ISSN 1514-6774, (2022).

[5] ISO/IEC 25010: Systems and software engineering - Systems and software Quality Requirements and Evaluation (SQuaRE) - System and software quality models, (2011).

[6] Olsina L.: Applicability of a Foundational Ontology to Semantically Enrich the Core and Domain Ontologies. In: 13th International Joint Conference on Knowledge Discovery, Knowledge Engineering and Knowledge Management, Vol. 2: KEOD (Knowledge Engineering and Ontology Development), Portugal, pp. 111-119, ISBN 978-989-758-533-3, (2021).

[7] Olsina L.: "Thing Foundational Ontology: ThingFO v1.3's Terms, Properties, Relationships and Axioms", Preprint in Research Gate, January 2022, Available at https://www.researchgate.net/publication/356789065_Thing_Foundational_Ontology_ThingFO_v13%27s_Terms_Properties_Relationships_and_Axioms, DOI: http://dx.doi.org/10.13140/RG.2.2.20945.20329. Also at https://doi.org/10.48550/arXiv.2107.09129.

[8] Olsina L., Tebes G., and Becker P.: "SituationCO v1.2's Terms, Properties, Relationships and Axioms -- A Core Ontology for Particular and Generic Situations ", Preprint in Research Gate, May 2021. Available at https://www.researchgate.net/publication/351858871_SituationCO_v12%27s_Terms_Properties_Relationships_and_Axioms_--_A_Core_Ontology_for_Particular_and_Generic_Situations. DOI: http://dx.doi.org/10.13140/ RG.2.2.23646.56644.

[9] Olsina L., Papa F., and Becker P.: "MetricsLDO v2.0's Terms, Properties, Relationships and Axioms -- A Metric-based Measurement Low-Domain Ontology", Preprint in Research Gate, November 2020, Available at https://www.researchgate.net/publication/345979051_MetricsLDO_v20's_Terms_Properties_Relationships_and_Axioms_--_A_Metric-based_Measurement_Low-Domain_Ontology, DOI: http://dx.doi.org/10.13140/RG.2.2.33998.69441.

[10] Olsina L., Papa F., and Becker P.: "IndicatorsLDO v2.0's Terms, Properties, Relationships and Axioms -- An Indicator-based Evaluation Low-Domain Ontology", Preprint in Research Gate, December 2020. Available at https://www.researchgate.net/publication/346717550_IndicatorsLDO_v20's_Terms_Properties_Relationships_and_Axioms_--_An_Indicator-based_Evaluation_Low-Domain_Ontology, DOI: http://dx.doi.org/10.13140/RG.2.2.27769.08809.

[11] Olsina L., Papa F., Molina H.: How to Measure and Evaluate Web Applications in a Consistent Way, *Web Engineering: Modelling and Implementing Web Applications*, Rossi G., Pastor O., Schwabe D., Olsina L. (Eds.), Springer HCIS, Chapter13, pp. 385-

420, (2008).

[12] Rivera M.B., Becker P., Olsina L.: Quality Views and Strategy Patterns for Evaluating and Improving Quality: Usability and User Experience Case Studies, In: *Journal of Web Engineering*, Rinton Press, US, 15:(5&6), pp. 433-464, ISSN 1540-9589, (2016).

[13] Tebes G., Olsina L., Peppino D., and Becker P.: "FRsTDO v1.1's Terms, Properties and Relationships -- A Top-Domain Functional Requirements Ontology", Preprint in Research Gate, September 2020, Available at https://www.researchgate.net/publication/344083773_FRsTDO_v11's_Terms_Properties_and_Relationships_--_A_Top-Domain_Functional_Requirements_Ontology, DOI: http://dx.doi.org/10.13140/RG.2.2.31659.26400.

[14] Tebes G., Olsina L., Peppino D., Becker P.: Specifying and Analyzing a Software Testing Ontology at the Top-Domain Ontological Level, *Journal of Computer Science & Technology*, 21(2), pp. 126-145, DOI: https://doi.org/10.24215/16666038.21.e12, (2021).

[15] Tebes G., Peppino D., Becker P., and Olsina L.: "TestTDO v1.3's Terms, Properties, Relationships and Axioms - A Top-Domain Software Testing Ontology", Preprint in Research Gate, August 2021, Available at https://www.researchgate.net/publication/353692577_TestTDO_v13%27s_Terms_Properties_Relationships_and_Axioms_-_A_Top-Domain_Software_Testing_Ontology, DOI: http://dx.doi.org/10.13140/RG.2.2.22622.92489. Also at https://doi.org/10.48550/arXiv.2104.09232.